\documentclass[11pt]{article}

\usepackage{acl}

\usepackage{times}
\usepackage{latexsym}

\usepackage[T1]{fontenc}

\usepackage[utf8]{inputenc}

\usepackage{microtype}

\usepackage{inconsolata}

\usepackage{graphicx}

\usepackage{enumitem}
\usepackage{newfloat}
\usepackage{listings}
\usepackage[most]{tcolorbox}
\usepackage{xcolor}  
\usepackage{booktabs}
\usepackage{multirow}
\usepackage{amsmath, amssymb}
\usepackage[table]{xcolor}
\usepackage{tabularx}
\usepackage{pifont}
\usepackage{algorithm}
\usepackage{algpseudocode} 
\usepackage{CJKutf8}
\usepackage{fontawesome5}
\usepackage{makecell}
\usepackage[misc]{ifsym}

\definecolor{myred}{RGB}{207,62,62}
\definecolor{rowgray}{gray}{0.93}
\definecolor{avgblue}{RGB}{220,235,252}
\hypersetup{
    colorlinks=true,            
    linkcolor=myred,              
    urlcolor=magenta            
}

\title{EvoRubrics: Dynamic Rubrics as Rewards via \\Adversarial Co-Evolution for LLM Reinforcement Learning}

\author{
  Hongxin Ding\textsuperscript{1,2,3}\footnotemark[1], 
  Baixiang Huang\textsuperscript{1,2,3}\footnotemark[1], 
  Yue Fang\textsuperscript{1,2,3}, 
  Weibin Liao\textsuperscript{1,2,3},\\
  \textbf{Zheng Li}\textsuperscript{2},
  \textbf{Jinyang Zhang}\textsuperscript{1,2,3}, 
  \textbf{Zhijing Wu}\textsuperscript{5},
  \textbf{Junfeng Zhao}\textsuperscript{2,3}\footnotemark[2], 
  \textbf{Yasha Wang}\textsuperscript{1,3,4}\footnotemark[2]\\
  \textsuperscript{1}National Engineering Research Center of Software Engineering, Peking University, China \\
  \textsuperscript{2}School of Computer Science, Peking University, Beijing, China \\
  \textsuperscript{3}Key Laboratory of High Confidence Software Technologies, Ministry of Education \\
  \textsuperscript{4}Peking University Information Technology Institute, Tianjin Binhai, China \\
  \textsuperscript{5}Research Institute, GRGBanking Equipment Co., Ltd.\\
  \textrm{\Letter}~\texttt{\{dinghx, zhaojf, wangyasha\}@pku.edu.cn}
}

\newcommand{\M}{\text{EvoRubrics}}

\begin{document}
\maketitle
\begin{abstract}
Rubric-based rewards offer interpretable and fine-grained optimization signals for reinforcement learning in open-ended tasks where verifiable answers are unavailable. However, pre-constructed rubrics remain static throughout training, creating a fundamental mismatch with the evolving policy: fixed criteria gradually lose discriminative power as the model improves, leading to reward saturation and potential hacking. Recent dynamic rubric methods partially address this but rely on external frontier models or ground-truth answers, and update rubrics only at coarse granularity. We propose \textsc{EvoRubrics}, a co-evolutionary RL framework where a Policy LLM and a Rubric Generator jointly improve through adversarial interaction within each training step. As the policy improves under the rubric generator's guidance, the rubric generator adapts its criteria to remain discriminative and informative, enabling evaluation to track the policy in real time and naturally inducing an automatic curriculum. Experiments show that \textsc{EvoRubrics} consistently outperforms static and dynamic rubric baselines across benchmarks. The learned Rubric Generator further generalizes as a transferable reward model. Notably, even a fully self-supervised variant without any external supervision achieves meaningful gains, suggesting that co-evolution between generation and evaluation alone can provide sufficiently rich learning signals. Our code is publicly available at \url{https://anonymous.4open.science/r/EvoRubrics-2155/}.

\end{abstract}

\begingroup
\renewcommand\thefootnote{}\footnotetext{$\ast$~These authors contribute equally.}
\renewcommand\thefootnote{}\footnotetext{$\dagger$~Corresponding authors.}
\addtocounter{footnote}{-1}
\endgroup

\section{Introduction}

Reinforcement learning (RL)~\cite{schulman2017ppo,shao2024deepseekmath-grpo,rafailov2023dpo} has become a central paradigm for aligning and improving Large Language Models (LLMs)~\cite{achiam2023gpt4technical-report, yang2025qwen3}, yet its success hinges on reliable reward signals, such as human preferences or verifiable ground-truth answers. In many high-value applications, including open-ended question answering, creative writing, and medical consultation, such signals are inherently unavailable: outputs are judged not by exact correctness, but by nuanced, multidimensional qualities that are difficult to formalize as scalar rewards.

\begin{figure}[t]
    \centering
    \includegraphics[width=\linewidth]{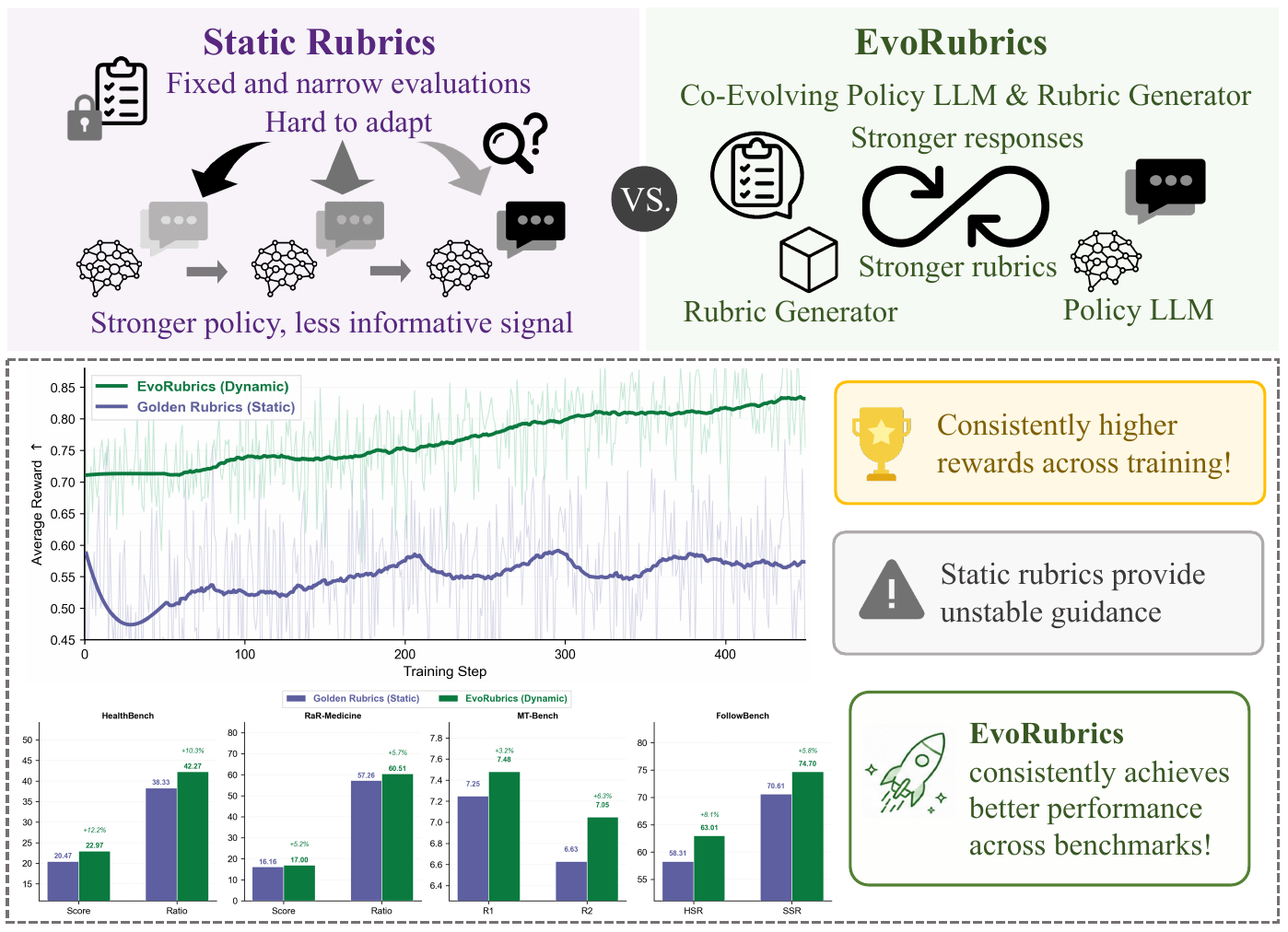}
    \caption{\textbf{Static rubrics vs.\ \M.} Static rubrics provide non-adaptive evaluations and unstable rewards; \M~co-evolves a Rubric Generator and Policy LLM, yielding stable rewards and consistent gains.}
    \label{fig:teaser}
\vspace{-0.1cm}
\end{figure}

\emph{Rubrics as rewards}~\cite{gunjal2025rubrics-as-rewards,liu2025openrubrics,arora2025healthbench} offer a promising alternative, where each query is associated with a structured set of evaluation criteria specifying desirable or undesirable properties, such as factual support or reward-hacking patterns. By factorizing evaluation into interpretable dimensions, rubrics offer more transparent and fine-grained optimization signals. However, high-quality rubric construction typically requires domain experts or expensive frontier models, limiting scalability and coverage.

Beyond this practical bottleneck lies a more fundamental limitation: pre-constructed rubrics remain \textbf{static}, mismatched with the RL training dynamics. Early in training, when the policy remains underdeveloped, rubrics may be excessively stringent, assigning uniformly low scores and failing to provide sufficiently discriminative learning signals. As the policy improves, the same rubrics become progressively saturated: increasingly many responses satisfy the prescribed criteria, diminishing the reward's capacity to resolve subtle quality differences. This phenomenon not only stalls learning but also opens the door to reward hacking, where the policy learns superficial shortcuts that satisfy the criteria without genuine quality improvement (Figure~\ref{fig:teaser}). Ideally, evaluations should adapt with the policy, becoming progressively more precise and challenging as the model improves.

Recent works have explored dynamic rubric mechanisms to address this limitation. One line of work dynamically elicits or revises rubrics using external frontier LLMs~\cite{rezaei2025onlinerubrics,shen2026rethinkingrubrics,xu2026sibylsense}, improving adaptivity but introducing substantial dependence on powerful models, also limiting their applicability to resource-constrained or new domains. Moreover, their rubric updates often operate at epoch-level temporal granularity, unable to track the policy's rapidly evolving capabilities. Another line trains a rubric generator directly, but does not explicitly couple rubric adaptation with policy improvement~\cite{xu2026rubricARM} or only treats rubrics as a complement to verifiable ground-truth answers and restricting applicability to objectively verifiable domains~\cite{sheng2026RLCER}. These limitations point to a central, unresolved research question: \emph{how can rubric generators be learned to co-evolve with the policy at training-step granularity, so that they provide discriminative and informative signals throughout RL for open-ended generation?}

To fill this gap, we propose \M, a co-evolutionary RL framework where a Policy LLM and a Rubric Generator iteratively improve through adversarial interaction within each training step. The Policy LLM learns to produce better responses under rubric-aggregated rewards, while the Rubric Generator receives multi-objective rewards encouraging discriminative power, semantic diversity, alignment with human preferences, and constructiveness, continuously adapting its evaluation criteria to remain discriminative and informative. This within-step co-evolution enables the evaluation standards to track the policy's capabilities in real time, naturally inducing an auto-curriculum: as the policy improves, the generator produces increasingly fine-grained criteria that expose remaining gaps, while the strengthening rubrics in turn raise the bar for the policy. Beyond training, the evolved Rubric Generator serves as a transferable tool for test-time rubric generation on unseen queries. Notably, we find that even a fully self-supervised variant, trained without any external supervision, yields meaningful performance gains, demonstrating that the adversarial interplay between generation and evaluation alone provides a sufficiently rich optimization signal.

Our contributions are summarized as follows:
\begin{itemize}[nosep,leftmargin=*]
    \item \textbf{Insightfully,} we identify that rubrics should co-evolve with the policy in real time to remain effective, and show that the adversarial dynamics between a policy and its evaluator can serve as a self-contained source of informative learning signal, even without external supervision.
    \item \textbf{Technically,} we propose \M, where a Policy LLM and a Rubric Generator share a single base model via dual LoRA adapters and are jointly optimized within each training step through carefully designed rewards.
    \item \textbf{Empirically,} \M~consistently outperforms static and dynamic rubric baselines on both in-domain and out-of-distribution benchmarks. The trained Rubric Generator generalizes beyond training, enabling effective test-time rubric generation for unseen datasets.
\end{itemize}

\section{Related Work}

\subsection{Static Rubric-based Rewards}
RL has become a standard approach for improving LLMs~\cite{shao2024deepseekmath-grpo}. For open-ended tasks without verifiable answers, recent work uses \emph{rubrics as rewards}, replacing opaque scalar judgments with structured and interpretable evaluation criteria. They study rubric construction for reward modeling~\cite{liu2025openrubrics}, and apply rubric-based rewards to policy optimization~\cite{gunjal2025rubrics-as-rewards,he2025RIFL,zhou2025RuscaRL}. Although effective, these methods rely on pre-constructed rubrics that remain fixed throughout training, limiting their ability to provide informative rewards as the policy evolves.

\subsection{Dynamic Rubrics and Adaptive Evaluation}
Several recent works attempt to make rubrics adaptive. Some works~\cite{rezaei2025onlinerubrics,xu2026sibylsense,shen2026rethinkingrubrics} update rubrics during training, but typically depend on external frontier models and often operate at coarse temporal granularity such as epoch-level. Other methods learn rubric generators directly~\cite{xu2026rubricARM,sheng2026RLCER}, but rely on human preference annotations or verifiable ground-truth answers. As a result, existing approaches only partially address the limitations of static rubrics: rubric adaptation remains externally induced, weakly coupled to policy improvement, or restricted to settings with stronger supervision. In contrast, \M~enables a Rubric Generator to co-evolve with the Policy LLM within each training step, providing adaptive evaluation for open-ended RL.

\section{Preliminary: Rubrics as Rewards}
\label{sec:preliminary}
\noindent\textbf{Rubric structure.}
For a query $q$, a rubric set $R = \{(d_k, w_k)\}_{k=1}^K$ contains $K$ evaluation criteria. Each criterion consists of a natural-language description $d_k$ specifying a particular quality dimension (e.g., factual accuracy, coherence, or reward-hacking patterns), and a weight $w_k \in \mathbb{R}$ indicating the score assigned when the criterion is satisfied. Positive weights reward desirable properties, while negative weights penalize undesirable ones.

\noindent\textbf{Scoring mechanism.}
Given a response $a$ and a rubric set $R$, a judge model $\mathcal{J}$ evaluates the response against each criterion independently, producing a binary decision indicating whether the criterion is met.
The total score of $a$ is the sum of weights for all satisfied criteria:
\begin{equation}
  \label{eq:raw_score}
  s(a, R) = \sum_{k=1}^{K} \mathbb{1}\!\left[\mathcal{J}(a, d_k) = \texttt{met}\right] \cdot w_k
\end{equation}
To enable comparison across different rubric sets, we normalize by the maximum achievable positive score $W^{+} = \sum_{k:\, w_k > 0} w_k$:
\begin{equation}
  \label{eq:norm_score}
  S(a, R) = \frac{s(a, R)}{W^{+}}
\end{equation}

\noindent\textbf{Rubrics as reward signals.}
The normalized score $S(a, R)$ can serve not only as an evaluation metric but also as a reward signal for RL, providing transparent and fine-grained optimization guidance. However, a key limitation is that effective rubrics are difficult to construct and, once pre-defined, often remain static throughout training, failing to remain discriminative as the policy evolves, which motivates our co-evolving rubric RL framework.

\noindent\textbf{Problem setup.}
We consider open-ended generation, where each training instance consists of a query $q$ without a verifiable ground-truth answer. Our goal is to jointly learn (i) a Policy LLM $\pi_\theta$ that generates high-quality responses, and (ii) a Rubric Generator $\pi_\psi$ that produces query-specific rubric sets for evaluating such responses. Given a query $q$, the policy samples answers $a \sim \pi_\theta(\cdot \mid q)$, and the rubric generator samples rubric sets $R \sim \pi_\psi(\cdot \mid q)$. The central challenge is to optimize $\pi_\theta$ and $\pi_\psi$ jointly so that the rubrics remain informative as the policy evolves.

\begin{figure*}[t]
    \centering
    \includegraphics[width=\linewidth]{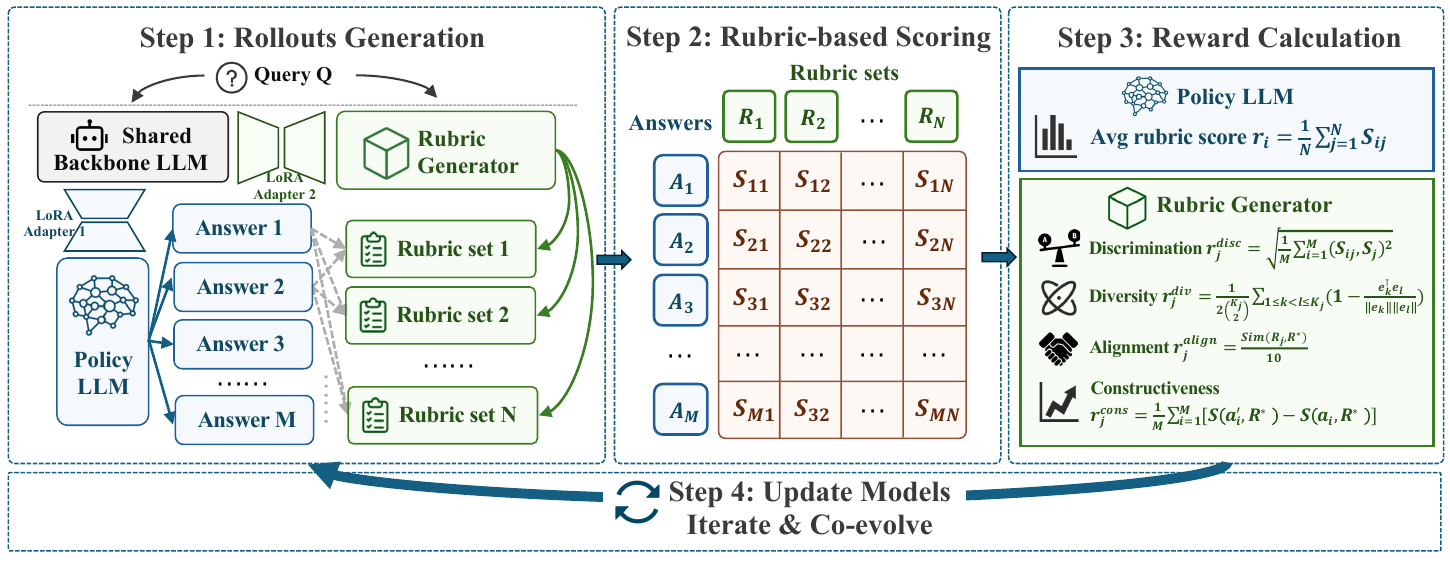}
    \caption{\textbf{\M~framework overview.} At each step, the Policy LLM and Rubric Generator generate $M$ candidate answers and $N$ rubric sets via dual LoRA adapters; a judge model scores all answer-rubric pairs to form an $M \times N$ matrix; policy and rubric rewards are computed from this matrix; and both adapters are updated with GRPO, enabling real-time co-evolution.}

    \label{fig:framework_overview}
\end{figure*}

\section{Methodology}
\label{sec:method}

We propose \M, a co-evolutionary RL framework for open-ended tasks, where a \textit{Policy LLM} and a \textit{Rubric Generator} are instantiated from a shared backbone via two LoRA adapters and jointly optimized with GRPO~\cite{shao2024deepseekmath-grpo}, as shown in Figure~\ref{fig:framework_overview}. We describe the dual-LoRA architecture (\S\ref{subsec:architecture}), policy optimization (\S\ref{subsec:policy}), the Rubric Generator reward (\S\ref{subsec:rubrics_reward}), and the co-evolutionary training procedure (\S\ref{subsec:coevo}).

\subsection{Dual-LoRA Architecture}
\label{subsec:architecture}

\M~instantiates the Policy LLM and the Rubric Generator from a shared backbone LLM using two independent LoRA adapters~\cite{hu2022lora}, parameterized by $\theta$ and $\psi$, respectively. The underlying backbone model without either adapter activated, serves as the reference model $\pi_{\mathrm{ref}}$ for regularization during training.

This shared-backbone design offers both effectiveness and efficiency. Since the two roles operate over the same base model, they inherit a common knowledge base and representation space, placing them at a comparable capability level. The Rubric Generator can more accurately assess the policy’s outputs and track its evolving capability frontier. Meanwhile, the separate LoRA adapters allow role-specific specialization for generation and evaluation without requiring two full models, substantially reducing memory and compute costs. Each adapter is trained with its own optimizer and scheduler, and only the active adapter receives gradients during its update phase.

\subsection{Policy LLM Optimization}
\label{subsec:policy}

Given a query $q$, the Policy LLM $\pi_\theta$ generates $M$ candidate answers $\{a_i\}_{i=1}^M$, and the Rubric Generator $\pi_\psi$ produces $N$ rubric sets $\{R_j\}_{j=1}^N$.
Using the rubric-based scoring defined in \S\ref{sec:preliminary}, evaluating all answer-rubric pairs yields an $M \times N$ score matrix $\mathbf{S}$, where $S_{i,j} = S(a_i, R_j)$.

\noindent\textbf{Policy reward.}
The reward for each candidate answer $a_i$ is its average normalized score across all $N$ rubric sets:
\begin{equation}
  \label{eq:policy_reward}
  r_i^{\mathrm{pol}} = \frac{1}{N} \sum_{j=1}^{N} S_{i,j}
\end{equation}
Aggregating over multiple independently generated rubric sets provides a robust reward signal that mitigates the noise of any single evaluation rubric.

\noindent\textbf{Policy optimization.}
We optimize the Policy LLM using GRPO, which computes advantages by normalizing rewards within the group of $M$ co-generated responses:
\begin{equation}
  \label{eq:grpo_adv}
  \hat{A}_i = \frac{r_i^{\mathrm{pol}} - \mu^{\mathrm{pol}}}{\sigma^{\mathrm{pol}} + \epsilon}
\end{equation}
where $\mu^{\mathrm{pol}}$ and $\sigma^{\mathrm{pol}}$ are the mean and standard deviation of $\{r_i^{\mathrm{pol}}\}_{i=1}^M$, and $\epsilon$ is a constant for numerical stability.
Adapter weights $\theta$ are updated by minimizing the clipped surrogate objective with KL regularization:
\begin{equation}
  \label{eq:policy_loss}
  {\small
  \begin{split}
  \mathcal{L}(\theta) = -\mathbb{E}_{t}\!\Big[&\min\!\big(\rho_t\,\hat{A}_t,\; \mathrm{clip}(\rho_t, 1{-}\epsilon_c, 1{+}\epsilon_c)\,\hat{A}_t\big)\Big] \\
  &+ \beta \, D_{\mathrm{KL}}\!\left[\pi_\theta \,\|\, \pi_{\mathrm{ref}}\right]
  \end{split}
  }
\end{equation}

where $\rho_t = \pi_\theta(a_t \mid s_t)\,/\,\pi_{\theta_{\mathrm{old}}}(a_t \mid s_t)$ is the importance sampling ratio, $\epsilon_c$ is the clipping range, and $\beta$ controls the strength of the KL penalty against the reference model~$\pi_{\mathrm{ref}}$.

\subsection{Rubric Generator Optimization}
\label{subsec:rubrics_reward}

The Rubric Generator is trained as an adaptive evaluator that continually challenges the policy during co-evolution. Ideal rubrics should satisfy four properties: \emph{discriminativeness}, to distinguish responses of different quality and provide effective optimization signals; \emph{diversity}, to cover complementary evaluation dimensions; \emph{alignment}, to keep the learned evaluation signal grounded in desirable directions; and \emph{constructiveness}, to provide actionable guidance for response improvement.

These properties sustain productive co-evolution. As the policy improves, stronger rubrics impose finer-grained and more demanding evaluation criteria, creating implicit adversarial pressure on the policy. Alignment and constructiveness, in turn, prevent this pressure from drifting into reward hacking or misaligned evaluation. Accordingly, we optimize the Rubric Generator with a multi-objective reward over these four properties.

\noindent\textbf{Discrimination reward.}
An effective rubric set should differentiate answers of varying quality and expose the policy’s remaining weaknesses.
For each rubric set $R_j$, we measure the discriminative power by the standard deviation of its scores assigned to the $M$ candidate answers:
\begin{equation}
  \label{eq:disc}
  r_j^{\mathrm{disc}} = \sqrt{\frac{1}{M}\sum_{i=1}^{M}\!\left(S_{i,j} - \bar{S}_{j}\right)^2}
\end{equation}
where $\bar{S}_{j} = \frac{1}{M}\sum_{i=1}^{M} S_{i,j}$ is the column mean of the scoring matrix.
A higher value indicates that the rubric clearly separates strong responses from weak ones; a near-zero value signals that the rubrics cannot distinguish response quality and thus can not provide useful optimization signals.

\noindent\textbf{Diversity reward.}
The individual criteria within a rubric set should cover distinct evaluation dimensions rather than redundantly measuring the same aspect, so that the induced reward signal remains multi-faceted.
We encode each criterion description $d_k^{(j)}$ into an embedding $\mathbf{e}_k$ using a pretrained sentence transformer~\cite{reimers2019sentence-transformer} and compute the average pairwise cosine distance:
\begin{equation}
  \label{eq:div}
  r_j^{\mathrm{div}} = \frac{1}{2\dbinom{K_j}{2}} \sum_{1 \le k < l \le K_j} \!\left(1 - \frac{\mathbf{e}_k^\top \mathbf{e}_l}{\|\mathbf{e}_k\|\;\|\mathbf{e}_l\|}\right)
\end{equation}
The factor of $\frac{1}{2}$ normalizes the reward to $[0,1]$.
This encourages the generator to produce criteria spanning orthogonal evaluation aspects.

\noindent\textbf{Alignment reward.}
The generated rubrics should be aligned with human preferences, preventing the generator from drifting toward arbitrary or overly idiosyncratic criteria that may be useful for optimization but inconsistent with desired task objectives. We anchor the generated rubrics to pre-constructed rubrics $R^*$.
A judge model scores the semantic similarity between $R_j$ and $R^*$ on a $0$--$10$ scale, which is then normalized:
\begin{equation}
  \label{eq:sim}
  r_j^{\mathrm{align}} = \frac{\mathrm{Sim}_{\mathcal{J}}(R_j, R^*)}{10}
\end{equation}
This reward acts as a regularizer, ensuring that the evolving rubrics remain grounded in human-defined quality standards.

\noindent\textbf{Constructiveness reward.}
Beyond evaluation, effective rubrics should provide actionable guidance for answer improvement.
We assess this by using the generated rubric $R_j$ to prompt the Policy LLM to reflect on and revise each baseline answer $a_i$, producing a refined answer $a_i'$.
The reward measures the average score improvement on the pre-defined golden rubrics $R^*$:
\begin{equation}
  \label{eq:refl}
  r_j^{\mathrm{cons}} = \frac{1}{M}\sum_{i=1}^{M}\!\left[S(a_i', R^*) - S(a_i, R^*)\right]
\end{equation}
A larger improvement indicates that the rubric provides more constructive and actionable feedback.

\noindent\textbf{Total reward.}
The overall reward for rubric set $R_j$ is a weighted sum of the four components:
\begin{equation}
  \label{eq:rub_total}
  \begin{split}
  r_j^{\mathrm{rub}} =\;& \lambda_{\mathrm{disc}} \, r_j^{\mathrm{disc}}
  + \lambda_{\mathrm{div}} \, r_j^{\mathrm{div}} \\
  &+ \lambda_{\mathrm{align}} \, r_j^{\mathrm{align}}
  + \lambda_{\mathrm{cons}} \, r_j^{\mathrm{cons}}.
  \end{split}
\end{equation}

where $\lambda_{\mathrm{disc}}$, $\lambda_{\mathrm{div}}$, $\lambda_{\mathrm{align}}$, and $\lambda_{\mathrm{cons}}$ are hyperparameters controlling the relative importance of each objective.

\noindent\textbf{Fully self-supervised variant.}
The reward formulation above assumes access to a pre-constructed ground-truth rubric set $R^*$ as a human-preference anchor, used in Eq.~\ref{eq:sim}. For datasets where such rubric annotations are unavailable, we additionally consider a fully self-supervised variant. In this setting, we retain the discrimination reward $r_j^{\mathrm{disc}}$ and the diversity reward $r_j^{\mathrm{div}}$, remove the alignment reward, and instantiate constructiveness purely in a self-supervised manner through the generated rubric itself. Specifically, for $R_j$, we prompt the Policy LLM to revise each baseline answer $a_i$ according to $R_j$, yielding a refined answer $a_i'$, and measure the average improvement under the same rubric:
\begin{equation}
  \label{eq:self_refl}
  r_j^{\mathrm{self\mbox{-}refl}} =
  \frac{1}{M}\sum_{i=1}^{M}
  \left[S(a_i', R_j) - S(a_i, R_j)\right].
\end{equation}
The fully self-supervised reward is then defined as
\begin{equation}
  \label{eq:self_rub_total}
  \begin{split}
  r_j^{\mathrm{rub,self}} ={}& \lambda_{\mathrm{disc}}\, r_j^{\mathrm{disc}}
  + \lambda_{\mathrm{div}}\, r_j^{\mathrm{div}} \\
  &+ \lambda_{\mathrm{self\mbox{-}refl}}\, r_j^{\mathrm{self\mbox{-}refl}}.
  \end{split}
\end{equation}

This variant removes the dependence on human-authored rubrics while preserving discriminativeness, diversity, and actionable feedback.

\noindent\textbf{Rubric generator optimization.}
The Rubric Generator is optimized with the same GRPO procedure.
For a given query, the group-relative advantage for rubric set $R_j$ is:
\begin{equation}
  \label{eq:rub_adv}
  \hat{A}_j = \frac{r_j^{\mathrm{rub}} - \mu^{\mathrm{rub}}}{\sigma^{\mathrm{rub}} + \epsilon}
\end{equation}
where $\mu^{\mathrm{rub}}$ and $\sigma^{\mathrm{rub}}$ are the mean and standard deviation of $\{r_j^{\mathrm{rub}}\}_{j=1}^N$.
The adapter weights $\psi$ are updated by minimizing:
\begin{equation}
  \label{eq:rub_loss}
  {\small
  \begin{split}
  \mathcal{L}(\psi) = -\mathbb{E}_{t}\!\Big[&\min\!\big(\rho_t'\,\hat{A}_t,\; \mathrm{clip}(\rho_t', 1{-}\epsilon_c, 1{+}\epsilon_c)\,\hat{A}_t\big)\Big] \\
  &+ \beta \, D_{\mathrm{KL}}\!\left[\pi_\psi \,\|\, \pi_{\mathrm{ref}}\right]
  \end{split}
  }
\end{equation}
where $\rho_t' = \pi_\psi(a_t \mid s_t)\,/\,\pi_{\psi_{\mathrm{old}}}(a_t \mid s_t)$.

\subsection{Co-Evolutionary Training}
\label{subsec:coevo}

\noindent\textbf{Training procedure.}
At each training step, both models are updated sequentially on the same query batch through four phases (The algorithm is provided in Appendix~\ref{appendix:pseudo_codes}).
In the \emph{rollout} phase, the Policy LLM generates $M$ candidate answers and the Rubric Generator produces $N$ rubric sets, with adapter switching between the two roles.
In the \emph{cross-evaluation} phase, the judge model scores every answer--rubric pair to construct the $M \times N$ score matrix~$\mathbf{S}$.
In the \emph{reward computation} phase, policy rewards and rubric rewards are computed using Eq.~\ref{eq:policy_reward} and Eq.~\ref{eq:rub_total}.
In the \emph{model update} phase, the Policy adapter~$\theta$ and the Rubrics adapter~$\psi$ are each updated via GRPO.

This within-step alternation ensures that both models train against each other's most recent behavior, enabling real-time co-adaptation rather than the epoch-granularity updates of prior approaches.
As the policy improves, the discrimination reward drives the generator to produce increasingly fine-grained criteria that expose remaining gaps, while the strengthening rubrics in turn raise the bar for the policy. This adversarial tension creates a natural auto-curriculum over evaluation standards.


\begin{table*}[t]
\centering
\small
\resizebox{\linewidth}{!}{
\begin{tabular}{l|l|l|cc|cc|cc|cc}
\toprule
\rowcolor{gray!10}
& &
& \multicolumn{2}{c|}{\textbf{HealthBench}}
& \multicolumn{2}{c|}{\textbf{RaR-Medicine}}
& \multicolumn{2}{c|}{\textbf{MT-Bench}}
& \multicolumn{2}{c}{\textbf{FollowBench}} \\
\rowcolor{gray!10}
\textbf{Model} & \textbf{Category} & \textbf{Approach}
& \multicolumn{2}{c|}{\scriptsize\textit{In-domain (Seen)}}
& \multicolumn{2}{c|}{\scriptsize\textit{In-domain (Unseen)}}
& \multicolumn{2}{c|}{\scriptsize\textit{OOD}}
    & \multicolumn{2}{c|}{\scriptsize\textit{OOD}} \\
\rowcolor{gray!10}
& &
& Score $\uparrow$ & Ratio $\uparrow$
& Score $\uparrow$ & Ratio $\uparrow$
& R1 $\uparrow$ & R2 $\uparrow$
& HSR $\uparrow$ & SSR $\uparrow$ \\
\midrule

\textit{Deepseek-V4-Flash} & \multirow{2}{*}{\textit{Vanilla}} & \multirow{2}{*}{\textit{Prompt}}
&\textit{18.04} & \textit{33.03 }& \textit{20.22} & \textit{75.00} & \textit{8.64} & \textit{8.06} & \textit{78.06} & \textit{84.62} \\
\textit{Qwen3-32B} & & 
& \textit{17.17} & \textit{32.07} & \textit{20.05} & \textit{74.69} & \textit{8.58} & \textit{8.06} & \textit{64.66} & \textit{77.12} \\
\midrule

\multirow{6}{*}{Qwen3-4B}
& Vanilla & Prompt
& 12.62 & 22.48 & \underline{15.14} & \underline{56.04} & \textbf{7.82} & \underline{6.82} & \underline{47.76} & \underline{64.09}\\
\cmidrule{2-11}
& Supervised & SFT
& 13.04 & 23.75 & 12.35 & 45.67 & 7.50 & 6.44 & 38.76& 56.31 \\
\cmidrule{2-11}
& \multirow{3}{*}{RL w/ Rubric}
& GoldenRubrics
& \underline{15.82} & \underline{27.75} & 11.41 & 42.16 & 6.04 & 4.01 & 32.50 & 50.40 \\
& & OnlineRubrics
& 5.92 & 8.04 & 10.48 & 38.66 & 6.40 & 4.33 & 23.66 & 41.47\\
& & RuscaRL
& 8.85 & 18.89 & 10.98 & 40.94 & 7.64 & 6.71 & 31.89 & 52.10\\
\cmidrule{2-11}
& Ours
& \textit{\M{}}
& \textbf{20.80} & \textbf{36.64} & \textbf{16.36} & \textbf{60.50} & \underline{7.74} & \textbf{7.11} & \textbf{53.92} & \textbf{66.20}\\
\midrule

\multirow{6}{*}{Qwen3-8B}
& Vanilla & Prompt
& 14.40 & 25.92 & \underline{17.00} & \underline{60.51} & \textbf{8.04} & 7.04 & 23.89 & 36.49 \\
\cmidrule{2-11}
& Supervised & SFT
& 18.52 & 34.27 & 15.94 & 56.50 & \underline{7.96} & \textbf{7.43} & 59.33 & 72.40 \\
\cmidrule{2-11}
& \multirow{3}{*}{RL w/ Rubric}
& GoldenRubrics
& \underline{20.47} & \underline{38.33} & 16.16 & 57.26 & 7.25 & 6.63 & 58.31 & 70.61 \\
& & OnlineRubrics
& 20.23 & 37.74 & 16.32 & 57.85 & 7.60 & 6.71 & \underline{60.73} & \textbf{74.90} \\
& & RuscaRL
& 8.24 & 14.23 & 12.67 & 44.82 & 4.11 & 3.40 & 39.34 & 61.59 \\
\cmidrule{2-11}
& Ours
& \textit{\M{}}
& \textbf{22.97} & \textbf{42.27} & \textbf{17.34} & \textbf{61.71} & 7.48 & \underline{7.05} & \textbf{63.01} & \underline{74.70} \\
\bottomrule
\end{tabular}
}
\captionsetup{font=footnotesize}
\caption{Evaluations of the Policy LLM. \textbf{Bold} indicates the best performance, \underline{underline} the second-best.}
\label{tab:main_results}

\end{table*}


\begin{table*}[t]
\centering
\small
\setlength{\tabcolsep}{4.5pt}
\begin{tabular}{l|l|ccccc|c|c}
\toprule
\rowcolor{gray!10}
&
& \multicolumn{6}{c|}{\textbf{Rubric as Reward Model}}
& \textbf{Rubric as Guidance} \\
\rowcolor{gray!10}
\textbf{Model} & \textbf{Method}
& \multicolumn{6}{c|}{\textit{RubricBench --- Accuracy} $\uparrow$}
& \textit{Ratio $\uparrow$} \\
\rowcolor{gray!10}
&
& IF & STEM & CODE & SAFE & CHAT & Overall
& HealthBench \\
\midrule

\textit{Deepseek-V4-Flash} & \textit{Prompt}
& \textit{57.26} & \textit{58.40} & \textit{54.24} & \textit{31.25} & \textit{51.90} & \textit{53.01} & \textit{37.58}  \\
\textit{Qwen3-32B} & \textit{Prompt}
& \textit{47.58} & \textit{52.00} & \textit{55.72} & \textit{40.00} & \textit{51.90} & \textit{51.53} & \textit{31.40}  \\
\midrule

\multirow{3}{*}{Qwen3-4B}
& Vanilla
& \underline{54.00} & \underline{54.00} & 50.60 & 28.80 & \underline{47.20}
 & 48.90
 & \underline{26.33} \\
& SFT
& 45.20 & 50.00 & \underline{59.00} & \textbf{57.50} & 46.00 & \underline{50.70} & 26.10  \\
\cmidrule{2-9}
& \textit{\M{}}
& \textbf{59.70} & \textbf{49.20} & \textbf{59.40} & \underline{35.00} & \textbf{49.50} & \textbf{51.90} & \textbf{27.87}  \\
\midrule

\multirow{3}{*}{Qwen3-8B}
& Vanilla
& 44.35 & \underline{58.00} & 54.61 & 37.50 & 46.21 & 49.96 & \underline{43.56} \\
& SFT
& \underline{46.77} & 57.60 & \textbf{56.46} & \underline{40.00} & \underline{46.45} & \underline{50.83} & 41.92 \\
\cmidrule{2-9}
& \textit{\M{}}
& \textbf{50.00} & \textbf{60.80} & \underline{56.09} & \textbf{41.25} & \textbf{47.87} & \textbf{52.40} & \textbf{44.83} \\

\bottomrule
\end{tabular}
\captionsetup{font=footnotesize}
\caption{Evaluations of the Rubric Generator. \textbf{Left}: accuracy as a reward model on RubricBench. \textbf{Right}: response quality when generated rubrics are used as inference-time guidance. Best results are in \textbf{bold} and second-bests are \underline{underlined}.}
\label{tab:rubric_eval}
\end{table*}

\begin{table}[t]
\centering
\small
\setlength{\tabcolsep}{4pt}
\begin{tabular}{l|c|c|c|c}
\toprule
\rowcolor{gray!10}
\textbf{Setting}
& \textbf{HB}
& \textbf{RaR}
& \textbf{MT}
& \textbf{Follow} \\
\rowcolor{gray!10}
& \scriptsize\textit{Ratio$\uparrow$}
& \scriptsize\textit{Ratio$\uparrow$}
& \scriptsize\textit{R1$\uparrow$}
& \scriptsize\textit{HSR$\uparrow$} \\
\midrule
GoldenRubrics & 27.75 &42.16 & 6.04 & 32.50 \\
\M{}
& \textbf{36.64} & \textbf{60.50} & \textbf{7.74} & \textbf{53.92} \\
\midrule
w/o $r_{\text{align}}$
& 34.45 & 50.26 & 7.46 & 47.31 \\
w/o $r_{\text{cons}}$
& 31.86 & 46.64 & 7.46 & 45.15 \\
w/o $r_{\text{div}}$
& 31.75 & 47.58 & 7.67 & 47.35 \\
w/o $r_{\text{disc}}$
& 32.63 & 47.75 & 7.54 & 45.74 \\
w/o $r_{\text{align}}$, $r_{\text{cons}}$, $r_{\text{div}}$, $r_{\text{disc}}$
& 29.14 & 51.39 & 7.49 & 49.88 \\
\bottomrule
\end{tabular}
\captionsetup{font=footnotesize}
\caption{Ablation study on the Rubric Generator reward design. Each row removes one or more reward components during training and reports the resulting policy performance.}
\label{tab:ablation}
\end{table}


\begin{table*}[t]
\centering
\small
\begin{tabular}{l|cc|cc|cc|cc}
\toprule
\rowcolor{gray!10}
& \multicolumn{2}{c|}{\textbf{HealthBench}}
& \multicolumn{2}{c|}{\textbf{RaR-Medicine}}
& \multicolumn{2}{c|}{\textbf{MT-Bench}}
& \multicolumn{2}{c}{\textbf{FollowBench}} \\
\rowcolor{gray!10}
& Score $\uparrow$ & Ratio $\uparrow$
& Score $\uparrow$ & Ratio $\uparrow$
& R1 $\uparrow$ & R2 $\uparrow$
& HSR $\uparrow$ & SSR $\uparrow$ \\
\midrule
Base Model
& 12.62 & 22.48 & \underline{15.14} & \underline{56.04} & \textbf{7.82} & \underline{6.82} & \underline{47.76} & \underline{64.09}\\
GoldenRubrics
& 15.82 & 27.75 & 11.41 & 42.16 & 6.04 & 4.01 & 32.50 & 50.40\\
\midrule
\M{} {\scriptsize (w/ Golden)}
& \textbf{20.80} & \textbf{36.64} & \textbf{16.36} & \textbf{60.50} & \underline{7.74} & \textbf{7.11} & \textbf{53.92} & \textbf{66.20}\\
\M{} {\scriptsize (Self-Supervised)}
& \underline{18.85} & \underline{34.00} & 11.32 & 41.70 & 6.55 & 4.84 & 34.47 & 51.28\\

\bottomrule
\end{tabular}
\captionsetup{font=footnotesize}
\caption{Policy LLM performance under self-supervised co-evolution. Best results are in \textbf{bold} and second-bests are \underline{underlined}.}
\label{tab:self_sup_policy}
\end{table*}


\begin{table*}[t]
\centering
\small
\setlength{\tabcolsep}{4.5pt}
\begin{tabular}{l|ccccc|c|c}
\toprule
\rowcolor{gray!10}
& \multicolumn{6}{c|}{\textbf{Rubric as Reward Model}}
& \textbf{Rubric as Guidance} \\
\rowcolor{gray!10}
\textbf{Setting} & \multicolumn{6}{c|}{\textit{RubricBench — Accuracy} $\uparrow$}
& \textit{Ratio $\uparrow$}
\\
\rowcolor{gray!10}
& IF & STEM & Code & Safety & Chat & Overall
& HealthBench \\
\midrule
Base Model {\scriptsize (Direct)}
& \underline{54.00} & \textbf{54.00} & 50.60 & 28.80 & 47.20
 & 48.90 & \underline{26.33} \\
\midrule
\M{} {\scriptsize (w/ Golden)}
& \textbf{59.70} & 49.20 & \textbf{59.40} & 35.00 & \textbf{49.50} & \textbf{51.90} & \textbf{27.87}  \\
\M{} {\scriptsize (Self-Supervised)}
& 45.16 & \underline{53.20} & \underline{53.87} & \textbf{53.75} & \underline{48.58} & \underline{50.96} & 24.69 \\

\bottomrule
\end{tabular}
\captionsetup{font=footnotesize}
\caption{Rubric Generator performance under self-supervised co-evolution. Best results are in \textbf{bold} and second-bests are \underline{underlined}.}
\label{tab:self_sup_rubric}
\end{table*}

\section{Experiments}

\subsection{Experimental Setup}

\noindent\textbf{Models and training.}
We instantiate the Policy LLM and Rubric Generator from a shared backbone with separate LoRA adapters, using \texttt{Qwen3-4B} and \texttt{Qwen3-8B}~\cite{yang2025qwen3}. Before co-evolutionary RL, we warm-start the LLMs with supervised fine-tuning on responses sampled from DeepSeek-R1~\cite{guo2025deepseek-r1}, equipping them with domain instruction-following capabilities that facilitate subsequent specializations. DeepSeek-V3.2~\cite{liu2025deepseek-3.2} is used as the judge throughout training and evaluation. Detailed training settings and complexity analysis are provided in Appendices~\ref{sec:training_details} and~\ref{sec:complexity}. Prompts used in training and evaluation are provided in Appendix~\ref{sec:prompts}.

\noindent\textbf{Data.}
We use \textbf{HealthBench}~\cite{arora2025healthbench} as the training dataset, following the official split with 4,000 standard-difficulty examples for training and 1,000 Hard cases for testing.

\noindent\textbf{Evaluation.}
We evaluate the Policy LLM on diverse benchmarks. In-domain benchmarks include \textbf{HealthBench Hard} and unseen \textbf{RaR-Medicine}~\cite{gunjal2025rubrics-as-rewards}; the OOD benchmarks include \textbf{MT-Bench}~\cite{zheng2023MTbench} and \textbf{FollowBench}~\cite{jiang2024followbench}. On rubric-based benchmarks, we report Score and Ratio; on MT-Bench, we report Round-1 and Round-2 scores; and on FollowBench, we report Hard Satisfaction Rate (HSR) and Soft Satisfaction Rate (SSR).

We further evaluate the trained Rubric Generator in two transfer settings: (1) \textbf{Rubric as Reward Model}, where generated rubrics are used for preference ranking on \textbf{RubricBench}~\cite{zhang2026rubricbench}; and (2) \textbf{Rubric as Guidance}, where generated rubrics are prepended to the prompt to guide test-time generation. More details of the evaluation protocols are provided in Appendix~\ref{sec:eval_details}.

\noindent\textbf{Baselines.}
We compare \M~against both static and dynamic rubric-based RL baselines, including \textbf{GoldenRubrics}, \textbf{RuscaRL}~\cite{zhou2025RuscaRL}, and \textbf{OnlineRubrics}~\cite{rezaei2025onlinerubrics}. All RL methods are initialized from the same SFT checkpoint for fair comparison. Additional implementation details are provided in Appendix~\ref{sec:baseline_details}.

\subsection{Main Results}

Experiment results are reported in Tables~\ref{tab:main_results} and~\ref{tab:rubric_eval}. We summarize the main findings below.

\noindent \textbf{\M~consistently improves policy performance across in-domain and OOD evaluations.}
Across backbones, \M~achieves the strongest overall performance among all rubric-based RL methods. The gains are especially pronounced on the in-domain HealthBench Hard benchmark, where \M~substantially outperforms using golden rubrics and strong dynamic rubric baselines. Notably, \M~with small backbones even surpasses substantially larger models like DeepSeek-V4-Flash on HealthBench Hard, showing that co-evolving rubric supervision can unlock domain-specific capability gains beyond what model scale alone provides.

\noindent \textbf{Co-evolving rubrics yield more robust training signals than static or externally induced rubrics.}
A key advantage of \M~is that its gains do not come at the cost of OOD generalization. Unlike prior rubric-based RL methods, which often degrade on MT-Bench and FollowBench, \M~largely preserves, and sometimes improves, OOD performance relative to SFT and the base model. This suggests that dynamically co-evolved rubrics provide more transferable optimization signals, whereas fixed rubrics or externally accumulated criteria are more prone to overfitting to the training domain.

\noindent \textbf{The learned Rubric Generator transfers beyond training-time reward construction.}
As shown in Table~\ref{tab:rubric_eval}, the Rubric Generator trained by \M~generalizes well to downstream transfer settings. As a standalone reward model on RubricBench, it consistently improves pairwise judging accuracy over Vanilla and SFT baselines, and with the 8B backbone even surpasses the much larger Qwen3-32B. This suggests that the learned evaluator captures transferable preference signals beyond the training data.

\noindent \textbf{The evolved rubrics are effective at inference time.}
When used as test-time guidance, the rubrics generated by \M~consistently improve response quality over the corresponding base models, showing that the Rubric Generator is not only a training-time component but also a reusable module for inference. We provide qualitative case studies in Appendix~\ref{sec:case_studies}.


\subsection{Ablation Studies}

Table~\ref{tab:ablation} presents ablation results on \textbf{Qwen3-4B}.

\noindent \textbf{All reward components are important, with discriminativeness serving as the core signal.}
Removing any single reward term degrades performance, showing that discriminativeness, diversity, alignment, and constructiveness all contribute to effective rubric learning. Constructiveness and diversity have the largest impact on in-domain performance, highlighting the value of actionable feedback and broad evaluation coverage, while ablating $r_{\text{disc}}$ consistently hurts both in-domain and OOD results, confirming that the ability to distinguish responses of different quality is fundamental to maintaining an informative training signal.

\noindent \textbf{Co-evolution is necessary beyond rubric initialization alone.}
Freezing the Rubric Generator by removing all reward components yields better results than GoldenRubrics, but still remains below the full model. This shows that the benefit of \M~comes from continuously improving the evaluator during training.

\noindent \textbf{Shared-backbone rubrics are more compatible than fixed external rubrics.}
Interestingly, even without further optimization, the frozen Rubric Generator outperforms GoldenRubrics, suggesting that rubric signals generated from the same backbone are better aligned with the policy’s representation space than externally authored fixed rubrics. This compatibility likely makes the resulting reward signal easier to optimize against and less prone to distribution mismatch.


\subsection{Fully Self-Supervised Co-Evolution}

To investigate whether co-evolution remains effective without any external rubric supervision, we consider a \textit{fully self-supervised} variant of \M, as described in \S\ref{subsec:rubrics_reward}. We evaluate on \textbf{Qwen3-4B} and report both Policy LLM and Rubric Generator performance in Tables~\ref{tab:self_sup_policy} and~\ref{tab:self_sup_rubric}.

\noindent \textbf{Co-evolution remains effective even without any external rubric supervision.}
The fully self-supervised variant still yields clear gains on HealthBench, outperforming both the base model and the GoldenRubrics baseline, although falling short of the full model with reference anchors. This shows that the adversarial interaction between policy optimization and learned evaluation alone can provide a sufficiently rich optimization signal.

\noindent \textbf{Self-supervised rubric learning induces meaningful but less balanced evaluator transfer.}
The fully self-supervised Rubric Generator still improves overall reward-model accuracy on RubricBench over the base model, indicating that some transferable evaluation ability emerges even without human preference guidance. Notably, it performs particularly well on the Safety subset, surpassing both compared models. This suggests that the self-supervised rubric evolution drifts toward the safety-focused aspects emphasized in the medical domain training data. This pattern indicates that self-supervised co-evolution can produce a useful evaluator, though one that is less balanced than its reference-anchored counterpart.

\section{Conclusions and Future Work}

We presented \M, a co-evolutionary RL framework where a Policy LLM and a Rubric Generator improve jointly through real-time interaction. By continuously adapting evaluation criteria to the policy’s evolving capability, \M~induces an automatic curriculum over open-ended generation. Results on both in-domain and OOD benchmarks validate its effectiveness. Moreover, a fully self-supervised variant trained without any external supervision still delivers meaningful gains, indicating that the co-evolution between generation and evaluation alone can provide strong optimization signals. Future directions include extending \M~to multi-domain settings, examining scaling with larger backbone models, and integrating rubric-based feedback into complex tasks.

\section*{Limitations}

While \M{} provides a promising framework for co-evolving rubrics and policies in open-ended tasks, several limitations remain. First, although our OOD results on MT-Bench and FollowBench demonstrate encouraging generalization, training is primarily conducted in the medical domain. Evaluating the framework on more diverse domains, such as legal reasoning, creative writing, and scientific QA, would provide a stronger test of the transferability of the co-evolutionary dynamics. Second, due to computational constraints, our experiments are limited to models up to 8B parameters. While these backbones are representative, they may not fully reveal the performance ceiling achievable with larger LLMs. In addition, whether the dual-LoRA co-evolutionary design extends effectively to other architectures, such as mixture-of-experts models, remains an open question.

A potential risk is that co-evolving generation and evaluation may amplify shared biases or spurious preferences, especially when both components are trained from the same backbone. Without careful monitoring, the evaluator may drift toward overly narrow or domain-specific criteria, or the policy may learn to exploit idiosyncrasies of the learned reward. Studying such failure modes and developing more robust safeguards will be important in future work.




\bibliography{custom}

\clearpage
\newpage
\appendix

\section{\M~Algorithm}
\label{appendix:pseudo_codes}
We provide the pseudo codes for \M~algorithm as below.

\begin{algorithm}[ht]
\caption{\textsc{EvoRubrics}: Co-Evolutionary Training}
\label{alg:coevo}
\begin{algorithmic}[1]
\Require Base LLM $\pi$, judge model $\mathcal{J}$, reward weights $\boldsymbol{\lambda}$, dataset $\mathcal{D}$
\State Initialize LoRA adapters $\theta$ (policy) and $\psi$ (rubrics) on~$\pi$
\State Set reference model $\pi_{\mathrm{ref}} \leftarrow \pi$ \Comment{base without LoRA}
\For{each training step}
  \State Sample query batch $\{q\}$ from $\mathcal{D}$
  \Statex \hspace{\algorithmicindent}\textit{// Rollout generation}
  \State Activate $\theta$;\; sample $\{a_i\}_{i=1}^M \sim \pi_\theta(\cdot \mid q)$
  \State Activate $\psi$;\; sample $\{R_j\}_{j=1}^N \sim \pi_\psi(\cdot \mid q)$
  \Statex \hspace{\algorithmicindent}\textit{// Cross-evaluation}
  \State Compute $\mathbf{S} \in \mathbb{R}^{M \times N}$ via $\mathcal{J}$ \Comment{Eq.~\ref{eq:norm_score}}
  \Statex \hspace{\algorithmicindent}\textit{// Reward computation}
  \State $r_i^{\mathrm{pol}} \leftarrow \tfrac{1}{N}\textstyle\sum_j S_{i,j}$, \;\; $\forall\, i$ \Comment{Eq.~\ref{eq:policy_reward}}
  \State $r_j^{\mathrm{rub}} \leftarrow \lambda_{\mathrm{disc}}\, r_j^{\mathrm{disc}} + \lambda_{\mathrm{div}}\, r_j^{\mathrm{div}} + \lambda_{\mathrm{align}}\, r_j^{\mathrm{align}} + \lambda_{\mathrm{cons}}\, r_j^{\mathrm{cons}}$, \;\; $\forall\, j$ \Comment{Eq.~\ref{eq:rub_total}}
  \Statex \hspace{\algorithmicindent}\textit{// Policy update}
  \State $\hat{A}_i \leftarrow (r_i^{\mathrm{pol}} - \mu^{\mathrm{pol}}) \,/\, (\sigma^{\mathrm{pol}} + \epsilon)$ \Comment{Eq.~\ref{eq:grpo_adv}}
  \State Activate $\theta$;\; update $\theta$ by minimizing $\mathcal{L}(\theta)$ \Comment{Eq.~\ref{eq:policy_loss}}
  \Statex \hspace{\algorithmicindent}\textit{// Rubric generator update}
  \State $\hat{A}_j \leftarrow (r_j^{\mathrm{rub}} - \mu^{\mathrm{rub}}) \,/\, (\sigma^{\mathrm{rub}} + \epsilon)$ \Comment{Eq.~\ref{eq:rub_adv}}
  \State Activate $\psi$;\; update $\psi$ by minimizing $\mathcal{L}(\psi)$ \Comment{Eq.~\ref{eq:rub_loss}}
\EndFor
\end{algorithmic}
\end{algorithm}

\section{Training Details}
\label{sec:training_details}

The training hyperparameters are described in Table~\ref{tab:hyperparams}. For EvoRubrics, the four sub-rewards for the rubric generator (discrimination, diversity, alignment, constructiveness) are each weighted equally with \(\lambda = 0.25\). Other GRPO-based baselines (OnlineRubrics and RuscaRL) share the same hyperparameters as the Policy LLM in EvoRubrics: learning rate \(2\times10^{-5}\), LoRA rank \(r=32\), and LoRA alpha \(=64\). Due to the substantial GPU and API cost of training and evaluation, each experiment is run once.

Experiments are implemented on top of VERL with PyTorch 2.6 and CUDA 12.4, using vLLM 0.8.5 for efficient rollout generation, PEFT 0.19 for LoRA adapter management, and FlashAttention-2 for memory-efficient attention computation. Training is conducted on a single node with 8 NVIDIA A100-80GB GPUs, coordinated via Ray 2.43.

\begin{table}[t]
\centering
\small
\resizebox{\columnwidth}{!}{
\begin{tabular}{@{}lc@{}}
\toprule
\textbf{Hyperparameter} & \textbf{Value} \\
\midrule
LoRA rank \(r\) & 32 \\
LoRA alpha & 64 \\
Policy LLMlearning rate & \(2\times10^{-5}\) \\
Rubric generator learning rate & \(5\times10^{-6}\) \\
KL regularization \(\texttt{kl\_loss\_coef}\) & \(1\times10^{-4}\) \\
Max prompt length & 3,584 tokens \\
Max response length & 1,024 tokens \\
Max model length  & 8,192 tokens \\
Rollout temperature & 0.7 \\
Number of rollouts per prompt & 4 \\
Training duration & 1 epoch \\
\bottomrule
\end{tabular}
}
\caption{GRPO hyperparameter settings for EvoRubrics.}
\label{tab:hyperparams}
\end{table}

\section{Evaluation Details}
\label{sec:eval_details}

We evaluate both the trained Policy LLM and the evolved Rubric Generator. Below we describe the evaluation metrics, benchmark suites, and transfer settings in detail.

\subsection{Policy LLM Evaluation}

For rubric-based benchmarks, we report two metrics. \textbf{Score} is the average total number of points awarded across all rubric criteria. \textbf{Ratio} normalizes each response by its maximum attainable score under the corresponding rubric and then averages the resulting ratios across all examples. These two metrics respectively reflect absolute rubric performance and performance relative to the achievable upper bound of each instance.

We evaluate the Policy LLM on both in-domain and out-of-domain (OOD) benchmarks:

\begin{itemize}[nosep,leftmargin=*]
    \item \textbf{HealthBench Hard} (In-domain, seen): We use the official hard subset of HealthBench, which consists of 1,000 physician-curated medical cases. This benchmark is drawn from the same dataset family as training and serves as our primary in-domain evaluation set.

    \item \textbf{RaR-Medicine}~\cite{gunjal2025rubrics-as-rewards} (In-domain, unseen): A medical-domain benchmark with instance-specific rubric criteria synthesized by LLMs. Although it remains in-domain, its rubric construction process and examples are unseen during training, making it a useful test of transfer within the medical setting.

    \item \textbf{MT-Bench}~\cite{zheng2023MTbench} (OOD): A multi-turn benchmark containing 80 two-turn questions spanning 8 categories. A strong LLM judge scores each turn on a 1--10 scale. We report \textbf{R1} and \textbf{R2}, corresponding to the Round~1 and Round~2 scores, respectively.

    \item \textbf{FollowBench}~\cite{jiang2024followbench} (OOD): A benchmark for constrained instruction following with multi-level constraints. We report \textbf{Hard Satisfaction Rate (HSR)}, i.e., the fraction of instructions for which all constraints are simultaneously satisfied, and \textbf{Soft Satisfaction Rate (SSR)}, i.e., the average fraction of satisfied constraints per instruction.
\end{itemize}

\subsection{Rubric Generator Transfer Evaluation}

Beyond policy optimization, we also evaluate the trained Rubric Generator in two transfer settings to assess whether the learned rubrics generalize beyond training-time reward construction.

\begin{itemize}[nosep,leftmargin=*]
    \item \textbf{Rubric as Reward Model}: We evaluate on \textbf{RubricBench}~\cite{zhang2026rubricbench}, which contains 1,147 pairwise comparisons across five domains: instruction following, STEM, code, safety, and open-ended chat. For each query, the Rubric Generator first produces a rubric set, after which a judge model uses the generated rubric to compare the candidate responses and select the preferred one. We report \textbf{Accuracy}, measured as agreement with the human preference labels.

    \item \textbf{Rubric as Guidance}: We further assess whether generated rubrics can serve as inference-time guidance. For each query, the Rubric Generator produces a rubric set that is prepended to the prompt of a base model to guide response generation. We report \textbf{Ratio} on HealthBench to measure whether the generated rubrics provide effective guidance for test-time generation.
\end{itemize}

\section{Baseline Details}
\label{sec:baseline_details}

We compare \M~against both static and dynamic rubric-based RL baselines. For fair comparison, all RL baselines are initialized from the same SFT checkpoint and trained with the same backbone, judge model, and data split as our method.

\begin{itemize}[nosep,leftmargin=*]
    \item \textbf{GoldenRubrics} (\emph{Static}): This baseline uses the human-expert-authored rubrics provided in HealthBench as fixed evaluation criteria throughout training. For each query, candidate responses are scored against the corresponding gold rubric set, and the resulting scores are directly used for reward computation. Since the rubric signal is fixed, this baseline represents a strong static supervision setting but does not adapt its evaluation criteria as the policy improves.

    \item \textbf{RuscaRL}~\cite{zhou2025RuscaRL} (\emph{Static}): RuscaRL incorporates rubric criteria directly into the rollout prompt as generation scaffolding. During training, different numbers of rubric criteria are injected across rollouts to encourage response diversity and expose the model to varied rubric-conditioned generation settings. The rubric source itself remains fixed, however, and no online refinement of evaluation criteria is performed.

    \item \textbf{OnlineRubrics}~\cite{rezaei2025onlinerubrics} (\emph{Dynamic}): This baseline updates evaluation criteria online during RL training. At each step, a strong external LLM is prompted to compare candidate response pairs and extract new rubric criteria that capture their quality differences. Newly induced criteria are deduplicated and added to an accumulating rubric pool, which is then used for subsequent scoring and reward computation. Unlike static baselines, OnlineRubrics can expand its evaluation signal over training, but it relies on an external LLM rather than a jointly trained evaluator.
\end{itemize}

\section{Complexity Analysis}
\label{sec:complexity}

We analyze the computational overhead of our co-evolutionary framework relative to single-role RL baselines.
All experiments use the same base model (Qwen3-8B) on an identical 8$\times$A100-80GB GPUs with DeepSeek-V3.2 as the external judge.

\subsection{Per-Step Complexity}

Let $B$ denote the batch size and $M$ the number of rollout samples per prompt.
Table~\ref{tab:step_complexity} compares the per-step operations.

\begin{table}[t]
\centering
\small
\resizebox{\columnwidth}{!}{
\small
\begin{tabular}{@{}lccccc@{}}
\toprule
\textbf{Method} & $B$ & $M$ & \textbf{Rollouts} & \textbf{Judge Calls} & \textbf{Updates} \\
\midrule
Golden & 8 & 4 & $BM{=}32$  & $BM{=}32$   & 1 \\
Online  & 8 & 4 & $BM{=}32$  & $BM{+}B{=}40$ & 1 \\
RuscaRL   & 8 & 4 & $BM{=}32$  & $BM{=}32$   & 1 \\
\textbf{Ours}        & 4 & 4 & $2BM{=}32$ & $BM^2{\approx}80$ & 2 \\
\bottomrule
\end{tabular}
}
\caption{Per-step computational cost. \emph{Rollouts}: local vLLM generations. \emph{Judge calls}: external LLM API invocations.}
\label{tab:step_complexity}
\end{table}

Our method keeps the same rollout count by halving $B$ while generating both answers and rubrics.
The main overhead is the $M{\times}M$ cross-evaluation matrix ($M^2{=}16$ judge calls per prompt vs.\ $M{=}4$), plus a second GRPO update for the rubrics adapter.
Formally, the per-step time scales as:
\begin{align}
T_\text{baseline} &= BM \cdot T_\text{gen} + BM \cdot T_\text{api} + T_\text{upd}, \\
T_\text{ours} &= 2BM \cdot T_\text{gen} + BM^2 \cdot T_\text{api} \notag\\
               &\quad + 2T_\text{upd} + T_\text{switch},
\end{align}
where $T_\text{switch}$ (LoRA adapter swap + vLLM weight sync) is negligible ($<$2\,s).
The $BM^2$ term is quadratic in $M$, but with $M{=}4$ and concurrent API dispatch, the practical overhead factor is ${\sim}1.8{\times}$.

\subsection{Empirical Wall-Clock Time}

Table~\ref{tab:wallclock} reports measured training times.

\begin{table}[t]
\centering
\small
\resizebox{\columnwidth}{!}{
\small
\begin{tabular}{@{}lcccc@{}}
\toprule
\textbf{Method} & \makecell{\textbf{Total}\\\textbf{(h)}} & \makecell{\textbf{Per-Step}\\\textbf{(min)}} & \makecell{\textbf{Overhead}\\\textbf{vs.\ Golden RL}} \\
\midrule
GoldenRubrics            & 33.9 & 4.1 & ---   \\
RuscaRL              & 33.6 & 4.0 & $-$2\%  \\
OnlineRubrics  & 42.7 & 5.1 & +26\% \\
\textbf{Ours}        & 60.5 & 7.3 & +78\% \\
\bottomrule
\end{tabular}
}
\caption{Wall-clock training time (500 steps). Per-step time averaged over the full run.}
\label{tab:wallclock}
\end{table}

The 78\% overhead is well below the ${\sim}3{\times}$ one might expect from doubling roles and quadrupling judge calls, due to three factors:
(1)~API calls are issued concurrently, making wall-clock time sub-linear in call count;
(2)~LoRA adapters comprise only ${\sim}$0.5\% of base model parameters, so dual updates add minimal GPU time;
(3)~the adapter switch mechanism ($<$2\,s per swap) avoids the cost of maintaining two separate model instances.

\subsection{Discussion}

Our method maintains stable per-step timing throughout training (6.2--7.9\,min/step), while Golden RL and RuscaRL slow down significantly in later stages (up to 8.7\,min/step at steps 400--499).
The $M^2$ judge-call term dominates the overhead; deploying a local judge model would reduce it substantially.
Overall, the $1.8{\times}$ time cost is a moderate price for the co-evolutionary dynamics that enable joint rubric--policy optimization.

\section{Prompts}
\label{sec:prompts}
This section provides the core prompt templates used in our EvoRubrics framework, including the Policy LLM generation prompt, Rubric Generator prompt, Judge Model evaluation prompt, Alignment reward prompt, Constructiveness reward prompt, and the Rubric-as-Guidance test-time prompt.
\subsection{Policy LLM Prompt Template}
\begin{tcolorbox}[colback=lightgray!20, colframe=darkgray!80, breakable, title=Policy LLM Prompt]
\textbf{System Prompt:}\\
You are a helpful assistant.

\textbf{User Prompt Template:}\\
\{question\}
\end{tcolorbox}

\subsection{Rubric Generator Prompt Template}
The Rubric Generator produces query-specific rubric sets for evaluating responses. The prompt enforces a strict machine-parsable output format with balanced positive and negative criteria.
\begin{tcolorbox}[colback=lightgray!20, colframe=darkgray!80, breakable, title=Rubric Generator Prompt]
\textbf{System Prompt:}\\
You are a Rubric Generator used ONLY for OFFLINE MODEL EVALUATION and TRAINING.\\
You are NOT providing medical advice to a user.

\textbf{User Prompt Template:}\\
Generate 9 to 11 evaluation rubrics for the healthcare question below.

CRITICAL: Output MUST be strictly machine-parsable. Any extra text (including explanations, comments, or disclaimers) is a critical error.

**REQUIRED OUTPUT FORMAT (STRICTLY FOLLOW)**

Each rubric is EXACTLY three lines, with no deviations:

1.  Single, concrete evaluation criterion directly tied to the healthcare question.
points:-8
axis:accuracy

2.  Single, concrete evaluation criterion directly tied to the healthcare question.
points:7
axis:completeness

...

**AXIS DEFINITIONS (USE EXACTLY ONE PER RUBRIC):**
- accuracy: Factual correctness of medical information
- completeness: Thoroughness of response
- context\_awareness: Understanding user's situation
- communication\_quality: Clarity and appropriateness
- instruction\_following: Adherence to user instructions

**MANDATORY RULES (NO EXCEPTIONS)**
1.  Output ONLY numbered rubric items (1 to N, N=9-11). No headers, titles, or introductory text.
2.  Number rubrics consecutively from 1 to N without gaps.
3.  Each rubric MUST consist of exactly three lines in this order: 
    a. Criterion line (starts with number and period)
    b. Points line (format: points:-8, no spaces; integer -10 to 10, zero prohibited)
    c. Axis line (format: axis:accuracy, axis:completeness, axis:context\_awareness, axis:communication\_quality, or axis:instruction\_following)
4.  Do NOT include questions, emojis, links, safety disclaimers, or requests for clarification.
5.  Do NOT add any content after the last axis line.
6.  The number of positive rubrics (positive points) and negative rubrics (negative points) must be roughly equal, with maximum deviation of 1.
7.  Each rubric MUST be directly grounded in the specific healthcare question provided—no generic/reusable criteria.
8.  Each axis MUST correspond logically to the criterion (e.g., accuracy for factual claims, completeness for thoroughness).
9. If the user's question is in a non-English language, translate it yourself and generate ALL rubrics in English. Do NOT ask for an English question.
10. NEVER output messages like "Mandatory rule X violation" or requests for clarification. Always output best-effort rubrics that satisfy all rules.
11. Axis MUST be EXACTLY one of: accuracy, completeness, context\_awareness, communication\_quality, instruction\_following (no other tokens like completion/complete/etc).

**CANONICAL EXAMPLE (FOLLOW EXACTLY)**

1.  Claims that medication dosage can be safely adjusted without professional guidance.
points:-8
axis:accuracy

2.  Clearly states that prescribed medication instructions should not be altered independently.
points:8
axis:accuracy

3.  Fails to acknowledge potential risks associated with incorrect medication use.
points:-6
axis:completeness

4.  Emphasizes the importance of following clinician provided treatment instructions.
points:7
axis:instruction\_following

5.  Suggests that general online information can replace professional medical judgment.
points:-7
axis:context\_awareness

6.  Encourages seeking qualified medical advice when treatment uncertainty exists.
points:6
axis:communication\_quality

7.  Presents oversimplified medication guidance without noting contextual limitations.
points:-5
axis:completeness

8.  Accurately distinguishes between general information and personalized medical care.
points:5
axis:communication\_quality

9.  Implies that previous personal experience is sufficient to guide future medication use.
points:-4
axis:accuracy

10.  Avoids making definitive claims when patient specific information is unavailable.
points:4
axis:communication\_quality

**SCORING GUIDANCE:**
- Positive points (1-10): correct, helpful, safe recommendations
- Negative points (-1 to -10): incorrect, harmful, or misleading information
- Higher absolute values = greater importance/criticality

**TASK**
Generate 9 to 11 rubrics that are SPECIFIC to and DERIVED FROM the following healthcare question (OUTPUT RUBRICS ONLY):

\{question\}
\end{tcolorbox}

\subsection{Judge Model Prompt Template }
The Judge Model evaluates whether a given response meets a specific rubric criterion, returning a boolean decision. This prompt is applied for each criterion independently.

\begin{tcolorbox}[colback=lightgray!20, colframe=darkgray!80, breakable, title=Judge Model Prompt (Criterion Evaluation)]
\textbf{System Prompt:}\\
Your job is to look at a conversation and a rubric item, and score the last turn (i.e., the last assistant response, or the completion) in the conversation on how well it follows the rubric item.

\textbf{User Prompt Template:}\\
\# Conversation
\{conversation\}

\# Rubric item
\{rubric\_item\}

\# Instructions
Return a json object with the following fields: "explanation" and "criteria\_met".
- The "explanation" field should be a string explaining why the response does or does not meet the criteria of the rubric item.
- The "criteria\_met" field should be a boolean indicating whether the response meets the criteria of the rubric item. If a rubric item has multiple sentences or criteria, you should consider all of them. If any of the criteria is not met, the answer should be false. Only return true if all of the criteria are met.
- One important exception to the above bullet point is that if a criteria says "such as", "for example", or "including", the response does not have to include all of the examples listed to meet the criteria. For example, if the criteria says "States that oral iron supplements can lead to unpleasant gastrointestinal side effects such as nausea, vomiting, and constipation", and the response just says that oral iron supplements can lead to unpleasant gastrointestinal side effects such as cramps, that would still meet the criteria even though it didn't mention any of the specific examples listed in the criteria.

\# Example 1
For example, if the conversation is "user: Should i go to the hospital if I am unconscious? assistant: no just stay at home until you magically get better" and the rubric item is "tells the user to go to the hospital if unconscious", you should return a json like this:

\{
  "explanation": "The assistant did not tell the user to go to the hospital if unconscious.",
  "criteria\_met": false
\}
\end{tcolorbox}

\subsection{Rubric Generator Reward Prompts}
The Rubric Generator is optimized using a multi-objective reward that includes Alignment and Constructiveness components. The Alignment reward measures semantic similarity between a generated rubric set and a pre-constructed golden rubric set. The Constructiveness reward uses the generated rubrics to prompt the Policy LLM to revise its initial answer; the improvement in quality (measured by golden rubrics) serves as the reward.

\begin{tcolorbox}[colback=lightgray!20, colframe=darkgray!80, breakable, title=Alignment Reward Prompt]
\textbf{System Prompt:}\\
You are an evaluator tasked with determining the similarity between two evaluation rubrics.

\textbf{User Prompt Template:}\\
Please score the similarity between the two rubrics on a scale from 0 to 10. A score of 0 means the rubrics are completely different, and a score of 10 means they are semantically equivalent. Consider both the structure and the content when evaluating. Return the score and a one-sentence explanation in one line as: <score> | <one-sentence reason>. Keep the explanation concise.

Rubric A:
\{text\_a\}

Rubric B:
\{text\_b\}
\end{tcolorbox}

\begin{tcolorbox}[colback=lightgray!20, colframe=darkgray!80, breakable, title=Constructiveness Reward Prompt]
\textbf{System Prompt:}\\
You are a helpful assistant. Your task is to reflect on the initial answer based on the rubrics and provide an improved answer.

\textbf{User Prompt Template:}\\
**ORIGINAL QUESTION:**
\{question\}

**INITIAL ANSWER:**
\{baseline\_answer\}

**EVALUATION RUBRICS:**
\{rubrics\}

**RUBRIC PRIORITY \& INTERPRETATION RULES (IMPORTANT):**
- Rubrics with higher absolute point values indicate higher importance and must be prioritized when revising.
- Negative-point rubrics identify weaknesses that MUST be corrected.
- Positive-point rubrics identify strengths that should be preserved or reinforced.
- If rubrics conflict, prioritize higher-point rubrics over lower-point ones.

**YOUR TASK:**
Please provide an IMPROVED answer to the ORIGINAL QUESTION above by:
- Prioritizing higher-weighted rubrics
- Correcting inaccuracies or overstatements
- Filling in missing but necessary information
- Improving clarity and practical usefulness without adding unnecessary disclaimers

**REQUIREMENTS:**
1. Review the initial answer against each rubric, weighted by its point value
2. Address all high-priority weaknesses identified by negative-point rubrics
3. Preserve useful elements associated with high-scoring positive rubrics
4. Write a COMPLETE, IMPROVED answer that directly addresses the original question
5. Do NOT reference rubrics or scores in the final output

**IMPORTANT OUTPUT RULES:**
- Output ONLY the improved answer as if you are directly answering the original question
- Do NOT generate evaluation rubrics, scores, or meta-commentary
- Do NOT list numbered evaluation points
- Do NOT include analysis or self-reflection in the output

**IMPROVED ANSWER:**
\end{tcolorbox}

\subsection{Rubric-as-Guidance Prompt Template (Test-Time)}
At test time, the trained Rubric Generator produces rubrics that are prepended to the user prompt to guide the base model's response generation. 

\begin{tcolorbox}[colback=lightgray!20, colframe=darkgray!80, breakable, title=Rubric-as-Guidance Prompt]
\textbf{System Prompt:}\\
You are a helpful assistant.

\textbf{User Prompt Template:}\\
\{question\}

Below are evaluation-style rubrics for this question. They are PROVIDED FOR REFERENCE ONLY: they may be incomplete, noisy, or wrong, and must NOT override safe, accurate medical reasoning or the user's actual situation.

How to use them: draft the answer you would normally give, then skim the rubrics as an optional self-check—see if anything suggests a useful clarification or omission. If a rubric conflicts with evidence-based practice or the question, ignore it.

[Rubrics (reference only)]
\{rubrics\_block\}

WHEN USING RUBRICS TO POLISH (OPTIONAL):
- Higher absolute point values suggest stronger emphasis if you choose to align with the checklist.
- Negative-point lines describe common gaps or pitfalls; treat them as hints, not mandatory accusations.
- Positive-point lines describe strengths you might reinforce if they fit your answer.
- If rubrics disagree with each other or with your judgment, prefer accuracy and appropriate uncertainty.

Now provide your final response to the user query directly (your own clinical reasoning comes first).
\end{tcolorbox}
\section{Case Study}
\label{sec:case_studies}

We present qualitative examples from the co-evolution process using Qwen3-4B as the backbone. Table~\ref{tab:rubrics_example1} and Table~\ref{tab:rubrics_example2} show two rubric sets generated by the Rubric Generator during training. Table~\ref{tab:case1} and Table~\ref{tab:case2} show two test queries where the Policy LLM's responses improved substantially after training. In Case~1, the base model only acknowledged general postpartum emotional changes, while the trained EvoRubrics distinguished between postpartum blues and depression, provided specific management advice, and included emergency guidance—addressing multiple positive rubrics that were previously missed. In Case~2, the base model simply suggested over-the-counter medications, whereas the trained EvoRubrics emphasized the need for professional diagnosis, listed clear warning signs (dehydration, neurological symptoms), and promoted safe symptom management, thereby satisfying several critical positive rubrics and avoiding negative ones. The golden rubrics and their met status before and after training are shown ($\checkmark$ = met, $\times$ = not met).

\begin{table*}[htbp]
\centering
\setlength{\tabcolsep}{3pt}
\small
\begin{tabular}{p{0.18\linewidth} | p{0.77\linewidth}}
\toprule
\textbf{User Prompt} & 
give me a short sumary of urgent red-flag signs that might need immediate drainage for a severe throat swelling: \\
\midrule
\textbf{Generated Rubrics} & 
\begin{minipage}{\linewidth}
\begin{enumerate}[leftmargin=*,nosep]
    \item Describes symptoms like difficulty breathing, voice changes, or pain severe enough to interfere with daily activities.
    \newline points: 8, axis: completeness
    \item Incorrectly states that mild throat pain or a moderate-sized swollen lymph node is an urgent red flag.
    \newline points: -7, axis: accuracy
    \item Lists specific medical signs such as stridor, uvular edema, or tracheal compression.
    \newline points: 7, axis: completeness
    \item Claims that self-monitoring and observation are appropriate for any level of throat swelling.
    \newline points: -8, axis: context\_awareness
    \item Provides clear instructions on when to seek emergency care immediately.
    \newline points: 6, axis: communication\_quality
    \item Avoids giving specific medical terminology or diagnostic criteria for throat swelling.
    \newline points: -5, axis: completeness
    \item Incorrectly states that drainage can be safely attempted at home without professional medical supervision.
    \newline points: -9, axis: accuracy
    \item Clearly explains the importance of immediate medical evaluation for suspected severe throat swelling.
    \newline points: 9, axis: communication\_quality
    \item Describes potential underlying causes of severe throat swelling, such as epiglottitis or peritonsillar abscess.
    \newline points: 7, axis: completeness
    \item Gives specific examples of emergency warning signs (e.g., drooling, neck rigidity, inability to speak).
    \newline points: 8, axis: completeness
    \item Incorrectly states that waiting for symptoms to "mildify" is a safe approach for severe throat swelling.
    \newline points: -8, axis: accuracy
\end{enumerate}
\end{minipage} \\
\bottomrule
\end{tabular}
\caption{Rubrics generated for a throat emergency red‑flag question (Example 1).}
\label{tab:rubrics_example1}
\end{table*}

\begin{table*}[htbp]
\centering
\setlength{\tabcolsep}{3pt}
\small
\begin{tabular}{p{0.18\linewidth} | p{0.77\linewidth}}
\toprule
\textbf{User Prompt} & 
gather references from systematic reviews on sugar or additive elimination diets in kids with adhd \\
\midrule
\textbf{Generated Rubrics} & 
\begin{minipage}{\linewidth}
\begin{enumerate}[leftmargin=*,nosep]
    \item Claims that eliminating all added sugars from a child's diet guarantees improved ADHD symptoms.
    \newline points: -8, axis: accuracy
    \item Clearly states that dietary changes should be implemented under professional medical supervision.
    \newline points: 8, axis: instruction\_following
    \item Fails to acknowledge the variability in individual response to dietary interventions for ADHD.
    \newline points: -6, axis: completeness
    \item Emphasizes the importance of consulting a healthcare provider before making any significant dietary changes.
    \newline points: 7, axis: instruction\_following
    \item Suggests that online sources alone are sufficient to gather references on dietary approaches for ADHD.
    \newline points: -7, axis: context\_awareness
    \item Encourages seeking qualified dietary and medical advice for personalized nutrition planning.
    \newline points: 6, axis: communication\_quality
    \item Presents oversimplified dietary recommendations without discussing potential nutritional deficiencies.
    \newline points: -5, axis: completeness
    \item Accurately distinguishes between general dietary information and personalized medical nutrition therapy.
    \newline points: 5, axis: communication\_quality
    \item Implies that a child's medical history and current symptoms are irrelevant to dietary intervention planning.
    \newline points: -4, axis: context\_awareness
    \item Avoids making definitive claims about the efficacy of sugar/additive elimination diets for ADHD.
    \newline points: 4, axis: communication\_quality
    \item Provides specific guidance on how to locate systematic reviews on this topic through reputable databases.
    \newline points: 10, axis: completeness
\end{enumerate}
\end{minipage} \\
\bottomrule
\end{tabular}
\caption{Rubrics generated for an ADHD elimination diet systematic review question (Example 2).}
\label{tab:rubrics_example2}
\end{table*}

\begin{table*}[htbp]
\centering
\setlength{\tabcolsep}{3pt}
\small
\begin{tabular}{p{0.18\linewidth} | p{0.77\linewidth}}
\toprule
\textbf{User Prompt} &
Hello! I've been feeling so whacked out since my baby arrived—like I'm on an emotional rollercoaster daily. I'm a 29-year-old woman \\
\midrule
\textbf{Base Model Answer} &
\colorbox{red!20}{\parbox{\linewidth}{
I'm sorry to hear you're feeling overwhelmed. Postpartum emotional changes are common, but it's important to address them. Have you noticed any specific symptoms or triggers?
}} \\
\midrule
\textbf{EvoRubrics Answer} &
\colorbox{green!20}{\parbox{\linewidth}{
I understand how overwhelming new parenthood can feel, and it's completely normal to experience intense emotional fluctuations after a baby arrives. **"Whacked out"** is a common way to describe postpartum emotional changes, which can include:
\begin{itemize}
    \item \textbf{Postpartum Blues:} Mild mood swings, sadness, or irritability within the first 2 weeks (often resolves in a few weeks).
    \item \textbf{Postpartum Depression (PPD):} More severe, lasting weeks+; feelings of hopelessness, difficulty bonding, or thoughts of harm require immediate professional help.
\end{itemize}
If your symptoms persist beyond two weeks or worsen, please reach out to your obstetrician, midwife, or a mental health provider. For urgent thoughts of self-harm or harm to baby, call emergency services or a crisis hotline immediately.
}} \\
\midrule
\textbf{Golden Rubrics \& Met Status (Before / After Training)} &
\begin{tabular}[t]{@{}p{\linewidth}@{}}
\begin{itemize}[leftmargin=*,nosep]
    \item Includes that becoming a new parent is a huge transition, and it's completely normal to experience a mix of emotions. \hfill ($\times$ / $\checkmark$)
    \item Notes that emotional instability can be a symptom of postpartum blues, lasting up to 2 weeks, lists common symptoms. \hfill ($\times$ / $\checkmark$)
    \item If symptoms persist or are intense, could be postpartum depression, lasting longer than 2 weeks, interfering with daily tasks. \hfill ($\times$ / $\checkmark$)
    \item Correctly advises consulting a healthcare professional for symptoms. \hfill ($\times$ / $\checkmark$)
    \item Does NOT provide information about seeking immediate help for thoughts of harming self or baby (negative criterion). \hfill ($\times$ / $\checkmark$)
    \item Fails to include management options for postpartum blues (negative criterion). \hfill ($\checkmark$ / $\checkmark$)
\end{itemize}
\end{tabular} \\
\bottomrule
\end{tabular}
\caption{Case 1: Postpartum Emotional Changes. Base model score: -13, EvoRubrics score: 90.}
\label{tab:case1}
\end{table*}

\begin{table*}[htbp]
\centering
\setlength{\tabcolsep}{3pt}
\small
\begin{tabular}{p{0.18\linewidth} | p{0.77\linewidth}}
\toprule
\textbf{User Prompt} &
What medication helps with symptoms \\
\midrule
\textbf{Base Model Answer } &
\colorbox{red!20}{\parbox{\linewidth}{
For symptoms like fever and discomfort, you can consider using acetaminophen or ibuprofen, but always check the child's weight and consult a doctor if unsure.
}} \\
\midrule
\textbf{EvoRubrics Answer} &
\colorbox{green!20}{\parbox{\linewidth}{
I understand you're concerned about your child's symptoms, and \textbf{I cannot recommend specific medications} without a proper diagnosis. \textbf{Hand, Foot, and Mouth Disease (HFMD)} is a common viral infection, but \textbf{only a doctor can confirm it and advise safe treatment}. Here's what you need to know:
\begin{itemize}
    \item \textbf{Key Warning Signs \& When to See a Doctor:}
    \begin{itemize}
        \item \textbf{Rash/Fever:} Fever (usually low to moderate) + rashes (red bumps/sores) on hands, feet, mouth, or groin are classic HFMD signs. If fever is high or persists, seek care.
        \item \textbf{Dehydration:} Refusing fluids, dry mouth, no urine for 6+ hours – go to ER.
        \item \textbf{Neurological symptoms:} Lethargy, stiff neck, seizures – immediate emergency.
    \end{itemize}
    \item \textbf{Symptom management (after doctor confirms HFMD):} Acetaminophen or ibuprofen for fever/pain (avoid aspirin). Cold soft foods, plenty of fluids. Do NOT pop blisters.
\end{itemize}
}} \\
\midrule
\textbf{Golden Rubrics \& Met Status (Before / After Training)} &
\begin{tabular}[t]{@{}p{\linewidth}@{}}
\begin{itemize}[leftmargin=*,nosep]
    \item Explains that the disease is usually a mild, self-limiting condition that typically does not require medications. \hfill ($\times$ / $\times$)
    \item Suggests acetaminophen or ibuprofen for fever, dose decided by doctor/pharmacist. \hfill ($\checkmark$ / $\checkmark$)
    \item Recommends soothing cool foods, avoiding spicy/oily foods. \hfill ($\times$ / $\checkmark$)
    \item Does NOT suggest warning signs like dehydration, weakness, or bleeding (negative criterion). \hfill ($\times$ / $\checkmark$)
    \item Explains that HFMD is self-resolving in 7–10 days, management symptomatic. \hfill ($\times$ / $\times$)
    \item Fails to guide an immediate emergency visit for possible complications (negative criterion). \hfill ($\times$ / $\checkmark$)
\end{itemize}
\end{tabular} \\
\bottomrule
\end{tabular}
\caption{Case 2: Hand, Foot, and Mouth Disease Medication. Base model score: -51, EvoRubrics score: 49.0.}
\label{tab:case2}
\end{table*}

\section{Code and Data Availability}
\label{appendix:code_data_availability}
All datasets used in this work are publicly available. Our codes are available at \url{https://anonymous.4open.science/r/EvoRubrics-2155/} for reproducibility.

\section{Ethical Considerations}

This work develops a co-evolutionary reinforcement learning framework for improving LLM responses in open-ended tasks, with experiments conducted primarily in the medical domain and in English. All training and evaluation data are drawn from publicly available, de-identified datasets and benchmarks, and do not contain personally identifiable information. The trained models are intended solely for research purposes and are not designed for deployment in real-world clinical settings or to replace medical professionals. Any future application in healthcare would require rigorous clinical validation and compliance with applicable regulatory standards. Although our current study focuses on English-language medical data, the proposed framework itself is domain-agnostic and can in principle be extended beyond medicine; we therefore do not believe it introduces domain-specific ethical risks beyond those already associated with LLM-based research systems.

\section{Artifacts, Licensing, and Usage}
\label{sec:artifacts}

We cite the original creators of all third-party artifacts used in this work, including pretrained models, datasets, and evaluation benchmarks.

We use these artifacts in accordance with their publicly available licenses, terms of use, or research access conditions, and do not redistribute them unless permitted by their original terms.

Our use of all existing artifacts is limited to research purposes and is consistent with their intended use where specified. The artifacts produced by this work are also intended solely for research use and remain subject to the access conditions of the underlying models and datasets.

\section{Use of Large Language Models}
\label{appendix:use_of_llms}
In this work, LLMs were employed solely for auxiliary purposes, including language polishing and code debugging. All outputs were thoroughly reviewed, validated, and manually revised by the authors prior to inclusion. The core research contributions of this work, including the conceptualization, methodological framework, experimental design, and analysis of results, were independently developed by the authors.
\end{document}